\documentclass[10pt,twoside]{article}
\usepackage[latin1]{inputenc}
\usepackage{amsmath}
\usepackage{dsfont}
\usepackage{graphicx}
\usepackage{yhmath}
\usepackage[table]{xcolor}
\usepackage{mathrsfs} % permet d'avoir le D cursive
\usepackage{amssymb}
\usepackage{url}
\usepackage{makecell}
\usepackage{arydshln}
\usepackage{multirow}
\usepackage{arydshln}
\usepackage{mathtools}
\usepackage{stmaryrd}  % pour intervalle entiers

\input epsf
\numberwithin{equation}{section}

\newtheorem{thm}{Theorem}[section]

\newtheorem{Def}[thm]{Definition\rm}

\newtheorem{rmrk}[thm]{Remark}

\newcommand{\E}{\ensuremath{\mathbb{E}}}
\newcommand{\R}{\ensuremath{\mathbb{R}}}
\newcommand{\Z}{\ensuremath{\mathbb{Z}}}

\newcommand{\N}{\ensuremath{\mathbb{N}}}

\newcommand{\cov}{\ensuremath{\mathrm{Cov}}}

\newcommand{\lip}{\ensuremath{\mathrm{Lip}}}
\definecolor{grisclair}{gray}{0.9}
\font\dsrom=dsrom10 scaled 1200
\def \ind{\textrm{\dsrom{1}}}
\DeclareMathOperator*{\argmin}{argmin}

\newcommand{\dx}{ {d_x} }

\newcommand{\dy}{ {d_y} }

\newcommand{\mk}{ { \mathcal{K}} }
\newcommand{\mx}{ { \mathcal{X}} }
\newcommand{\my}{ { \mathcal{Y}} }
\newcommand{\mz}{ { \mathcal{Z}} }

\newcommand{\nm}[1]{ \|#1\| }

\newcommand{\qed}{\hfill $\blacksquare$}

\begin{document}
%\date{ }
\title{\bf Excess risk bound for deep learning under weak dependence}
 \maketitle \vspace{-1.0cm}
\begin{center} 
     William KENGNE \footnote{Developed within the ANR BREAKRISK: ANR-17-CE26-0001-01, the MME-DII center of excellence (ANR-11-LABEX-0023-01) and the  CY Initiative of Excellence (grant "Investissements d'Avenir" ANR-16-IDEX-0008), Project "EcoDep" PSI-AAP2020-0000000013} 
 \end{center}

  \begin{center}
  { \it 
 THEMA, CY Cergy Paris Université, 33 Boulevard du Port, 95011 Cergy-Pontoise Cedex, France\\
 % E-mail: william.kengne@u-cergy.fr\\
 % $^{\text{b}}$   LERSTAD, Université Gaston Berger, Saint-Louis, Sénégal. \\
  E-mail: william.kengne@cyu.fr  \\
  }
\end{center}

 \pagestyle{myheadings}
 \markboth{Excess risk bound for deep learning under weak dependence}{W. Kengne}

~~\\
\textbf{Abstract}:
This paper considers deep neural networks for learning weakly dependent processes in a general framework that includes, for instance, regression estimation, time series prediction, time series classification.
The $\psi$-weak dependence structure considered is quite large and covers other conditions such as mixing, association,$\ldots$ 
Firstly, the approximation of smooth functions by deep neural networks with a broad class of activation functions is considered.  
 We derive the required depth, width and sparsity of a deep neural network to approximate any H\"{o}lder smooth function, defined on any compact set $\mx$.
Secondly, we establish a bound of the excess risk for the learning of weakly dependent observations by deep neural networks.
When the target function is sufficiently smooth, this bound is close to the usual $\mathcal{O}(n^{-1/2})$.  
 
 \medskip
 

\section{Introduction}\label{sect_intro}
We focus on the supervised learning framework and consider the training sample $D_n=\{Z_1=(X_1,Y_1),\cdots, Z_n=(X_n,Y_n) \}$ which is a trajectory  of a stationary and ergodic process $\{Z_t=(X_t,Y_t), ~ t \in \Z \}$, taking values in $ \mathcal{Z} = \mathcal{X} \times \mathcal{Y}$, where $\mathcal{X}$ is the input space and $\mathcal{Y}$ the output space. 
In the sequel, we assume that $\mathcal{X} \subset \R^{d_x}$ and  $\mathcal{Y} \subset \R^{d_y}$, with $d_x, d_y \in \N$.
Denote by $ \mathcal{F}(\mathcal{X}, \mathcal{Y}) $ the set of measurable functions from $\mathcal{X}$ to $\mathcal{Y}$.
Consider a loss function $\ell : \R^{d_y} \times \mathcal{Y} \rightarrow [0,\infty)$ and for a predictor $h \in \mathcal{F}(\mathcal{X}, \mathcal{Y})$, define the risk,
\begin{equation}\label{def_risk}
  R(h) = \E_{Z_0}\big[\ell \big(h(X_0),Y_0 \big) \big] .
   \end{equation}
For a learner $h \in \mathcal{F}(\mathcal{X}, \mathcal{Y})$, a prediction of $Y_t$ is $\widehat{Y}_t = h(X_t)$ for all $t \in \Z$. 
The "best" predictor $h^* \in \mathcal{F}(\mathcal{X}, \mathcal{Y})$, when it exists, is the one that achieves the smallest risk, that is,
\begin{equation}\label{cond_best_pred}
  R(h^*) =  \underset{h \in \mathcal{F}(\mathcal{X}, \mathcal{Y})}{\inf} R(h) .
  \end{equation}
We would like to construct a learner $h \in \mathcal{F}(\mathcal{X}, \mathcal{Y})$ such that, for any $t\in \Z$, $h(X_t)$ is average "close" to $Y_t$; that is, a learner with the smallest risk.
But in general, this risk cannot be minimized in practice, because the distribution of $(X_0,Y_0)$ is unknown.
 Consider the empirical risk, defined for all $h \in \mathcal{F}(\mathcal{X}, \mathcal{Y})$ by,
\begin{equation}\label{def_emp_risk}
 \widehat{R}_n(h) = \frac{1}{n} \sum_{i=1}^n \ell\big(h(X_i),Y_i \big).
 \end{equation}
Thus, the aim is to build from the training sample $D_n$, a learner $\widehat{h}_n$, that minimizes the empirical risk.
In the sequel, we set $\ell(h,z) = \ell\big(h(x),y\big)$ for all $z=(x,y) \in  \mathcal{X} \times\mathcal{Y}$ and $h \in \mathcal{F}(\mathcal{X}, \mathcal{Y})$.
\medskip

We focus on the class of deep neural networks (DNN) predictors, with a general activation function $\sigma : \R \rightarrow \R$. 
Recall that, a neural network architecture $(L, \mathbf{p})$ stands for a positive integer $L$ called the number of hidden layers or depth and a width vector $\mathbf{p} = (p_0,p_1,\cdots,p_{L+1}) \in \N^{L+2}$. Therefore, a DNN with network architecture $(L, \mathbf{p})$ is any function of the
form,

\begin{equation} \label{DNN_def}
 h : \R^{p_0} \rightarrow \R^{p_{L+1}}, ~ x \mapsto h(x) = A_{L+1} \circ \sigma_L \circ A_L \circ \sigma_{L-1} \circ A_{L-1}\circ \cdots \circ \sigma_1 \circ A_1(x),
\end{equation} 
where for any $\ell = 1,\cdots, L+1$, $A_{\ell}: \R^{p_{\ell-1}} \rightarrow \R^{p_{\ell}}$ is an affine linear application defined by $A_\ell (x) := W_\ell x + b_\ell$ for given $p_{\ell-1} \times p_\ell$ weight matrix $W_\ell$, a shift vector $b_\ell \in \R^{p_\ell}$ and $\sigma_\ell : \R^{p_\ell} \rightarrow \R^{p_\ell}$ is an element-wise nonlinear activation map defined as $\sigma_{\ell}(z)=(\sigma(z_1),\cdots,\sigma(z_{p_\ell}) )^T $ for all $z=(z_1,\cdots,z_{p_\ell})^T \in \R^{p_\ell}$, and $^T$ denotes the transpose. 
In the setting considered here, $p_0 = d_x$ (input dimension) and $p_{L+1} = d_y$ (output dimension).
For a DNN $f$ of the form (\ref{DNN_def}), the vector of its parameters is denoted by $\theta(h)$, that is,
\begin{equation*}
\theta(h) = \big( \text{vec}(W_1)^T, b_1^T,\cdots, \text{vec}(W_{L+1})^T, b_{L+1}^T \big),
\end{equation*}
where $\text{vec}(W)$ denotes the vector obtained by concatenating the column vectors of the matrix $W$.
In the sequel, we deal with an activation function $\sigma$ and denote by 
$\mathcal{H}_{\sigma,d_x,d_y}$ the set of DNNs with $d_x$ dimensional input and $d_y$ dimensional output.
For any $h \in \mathcal{H}_{\sigma,d_x,d_y}$ with a neural network architecture $(L, \mathbf{p})$, $\text{depth}(h)$ denotes its depth and $\text{width}(h)$ denotes its width, that is, $\text{depth}(h) = L$ and $\text{width}(h) = \underset{1\leq \ell \leq L}{\max} p_\ell$. 
For any $L, N, S, B, F>0$, consider the sets,
\begin{equation*}
\mathcal{H}_{\sigma,\dx,\dy} (L,N) = \big\{ h \in \mathcal{H}_{\sigma,d_x,d_y}, ~  \text{depth}(h) \leq L, ~\text{width}(h) \leq N \big\},
\end{equation*}
\begin{equation}\label{def_H_lnsb}
\mathcal{H}_{\sigma,\dx,\dy} (L,N, S, B) = \big\{ h \in \mathcal{H}_{\sigma,\dx,\dy} (L,N), ~  |\theta(h)|_0 \leq S, ~\|\theta(h) \| \leq B \big\},
\end{equation}
\begin{equation}\label{def_H_lnsbf}
\mathcal{H}_{\sigma,\dx,\dy} (L,N, S, B, F) = \big\{ h \ind_{\mx}, ~ h \in \mathcal{H}_{\sigma,\dx,\dy} (L,N,S,B), ~\|h\|_{\infty,\mx} \leq F \big\},
\end{equation}
where for any $x=(x_1,\cdots,x_d)^T \in \R^d$, $|x|_0=\sum_{i=1}^d \ind(x_i \neq 0) $, $\|x\| = \underset{1\leq i \leq d}{\max} |x_i|$ and $\|h\|_{\infty,\mx} = \sup_{x \in \mx} \|h(x)\| $ (see also the Subsection \ref{sub_sec_assump} bellow).

\medskip

There are many works in the literature on the approximation of smooth functions by DNNs. We refer to \cite{yarotsky2017error}, \cite{suzuki2018adaptivity}, \cite{petersen2018optimal}, \cite{ohn2019smooth}, \cite{schmidt2019deep}, \cite{schmidt2020nonparametric} and the references therein for an
overview.
The existing results are mainly obtained, either for a restricted class of activation function (for example, ReLU), or for a specific case of input space (for example, $\mx=[0,1]^\dx$).
The first contribution of this paper:
\begin{itemize}
\item[(i)] Establish approximation results by DNNs with a quite large class of activation function, of smooth functions defined on any compact input space $\mx \subset \R^\dx$. 
\end{itemize}

\medskip

The second issue of this work is the learning of the process $\{Z_t=(X_t,Y_t), ~ t \in \Z \}$ by the class of DNNs $\mathcal{H}_{\sigma,\dx,\dy} (L,N, S, B, F)$ with $L,N, S, B, F >0$.
 There is a large literature that addresses such question with independent and identically distributed (i.i.d.) observations. See, among others papers, \cite{bauer2019deep}, \cite{ohn2019smooth}, \cite{schmidt2020nonparametric}, \cite{kim2021fast}, \cite{ohn2022nonconvex}. 
 For some theoretical results on dependent observations, see for instance, \cite{kohler2020rate}, \cite{kurisu2022adaptive}, \cite{ma2022theoretical}.
 These works are for mixing observations and/or for the regression problem.
 The recent work of \cite{kengne2023deep} considers a general framework of supervised learning for weakly dependent processes, where the weak dependence structure covers mixing conditions. But, the convergence of the excess risk is not addressed.
The second contribution of this paper:
\begin{itemize}
\item[(ii)] Derive the convergence rate of the excess risk for the learning of $\psi$-weakly dependent processes by DNNs.
This rate is close to the usual $\mathcal{O}(n^{-1/2})$ when the target function is sufficiently smooth. 
\end{itemize}

\medskip

The rest of the paper is structured as follows. Section 2 sets some notations and assumptions. Section 3 focuses on H{\"o}lder smooth functions approximation by DNNs. Section 4 considers the excess risk and provides its convergence rate, whereas Section 5 is
devoted to the proofs of the main results.

\section{Notations and assumptions}

\subsection{Some notations}\label{sub_sec_assump}
Throughout the sequel, the following notations will be used, with $d \in \N$, and where $E_1, E_2$ are subsets of separable Banach spaces equipped with norms $\| \cdot\|_{E_1}$ and $\| \cdot\|_{E_2}$ respectively.
\begin{itemize}
\item $\N_0 = \N \cup \{0\}$.
\item For all $x \in \R$, $[x]$ denotes the integer part of $x$, $\lceil x \rceil$ denotes the  smallest integer $\geq x$ and $\lfloor x \rfloor$ denotes the largest integer $\leq x$.
\item  $\|x\| = \underset{1\leq i \leq d}{\max} |x_i|$,  $|x|_0=\sum_{i=1}^d \ind(x_i \neq 0) $ for all $x=(x_1,\cdots,x_d)^T \in \R^d$.
\item  $ \|x \| = \underset{1\leq i \leq p}{\max} \sum_{j=1}^{q} |x_{i,j}| $ for any matrix $x=(x_{i,j}) \in M_{p,q}(\R)$; where $M_{p,q}(\R)$ denotes the set of matrices of dimension $p\times q$ with coefficients in $\R$.
\item For any function $h: E_1 \rightarrow E_2$ and $U \subseteq E_1$,
\[ \| h\|_\infty = \sup_{x \in E_1} \| h(x) \|_{E_2}, ~ \| h\|_{\infty,U} = \sup_{x \in U} \| h(x) \|_{E_2} \text{ and }\] 
 \[\lip_\alpha (h) \coloneqq \underset{x_1, x_2 \in E_1, ~ x_1\neq x_2}{\sup} \dfrac{\|h(x_1) - h(x_2)\|_{E_2}}{\| x_1- x_2 \|^\alpha_{E_1}} 
 \text{ for any } \alpha \in [0,1] .\]
 \item For any $\mathcal{K}>0$ and $\alpha \in [0,1]$, $\Lambda_{\alpha,\mathcal{K}}(E_1,E_2)$ (simply $\Lambda_{\alpha,\mathcal{K}}(E_1)$ when $E_2 \subseteq \R$) denotes the set of functions  $h:E_1^u \rightarrow E_2$ for some $u \in \N$, such that  $\|h\|_\infty < \infty$ and  $\lip_\alpha(h) \leq \mathcal{K}$.
When $\alpha=1$, we set  $\lip_1 (h)=\lip(h)$ and $\Lambda_{1}(E_1) =\Lambda_{1,1}(E_1,\R)$. 
\item  $\mathcal{F}(E_1, E_2)$ denotes the set of measurable functions from $E_1$ to $E_2$.
\item For any $h \in \mathcal{F}(E_1, E_2)$ and $\epsilon >0$, $B(h,\epsilon)$ denotes the ball of radius $\epsilon$ of $\mathcal{F}(E_1, E_2)$ centered at $h$, that is,
$B(h,\epsilon) = \big\{ f \in \mathcal{F}(E_1, E_2), ~ \| f - h\|_\infty \leq \epsilon \big\}$.
\item For any $\mathcal{H} \subset \mathcal{F}(E_1, E_2)$, the $\epsilon$-covering number $\mathcal{N}(\mathcal{H},\epsilon)$ of $\mathcal{H}$ is the minimal number of balls of radius $\epsilon$ needed to cover  $\mathcal{H}$; that is,
\[ \mathcal{N}(\mathcal{H},\epsilon)= \inf\Big\{ m \geq 1 ~ : \exists h_1, \cdots, h_m \in \mathcal{H} \text{ such that } \mathcal{H} \subset \bigcup_{i=1}^m B(h_i,\epsilon)    \Big\} .\]
\item For any $U \subset E_1$, $\overline{U}$ denotes the closure of $U$.
\item For any function $g: \R \rightarrow \R$, $g'$ and $g''$ denote the first and second order derivatives of $g$. Also, for all $x \in \R$, we set 
$g'(x+):= \lim_{\epsilon \downarrow 0} \big( g(x+\epsilon) - g(x) \big)/\epsilon$ and $g'(x-):= \lim_{\epsilon \downarrow 0} \big( g(x-\epsilon) - g(x) \big)/\epsilon$.
%\item For all $x \in \R$, $\lceil x \rceil$ denotes the  smallest integer $\geq x$ and $\lfloor x \rfloor$ denotes the largest integer $\leq x$.
\item For any bounded set $U \subset E_1$, $\| U \| = \sup_{x \in U} \|x\|$.
\item For all $x=(x_1,\cdots,x_d) \in \R^d$ and $\beta=(\beta_1,\cdots,\beta_d) \in \N_0^d$, $x^\beta=x_1^{\beta_1}\cdot \ldots \cdot x_d^{\beta_d}$ and $\beta! = \beta_1!\cdot\ldots\cdot\beta_d!$. 
\end{itemize}
% 
 % \medskip 

\subsection{Some assumptions and weak dependence}
 Let us set the following assumptions on the input space $\mathcal{X}$ and the loss function $\ell$.
 \begin{enumerate} 
%     \item [(\textbf{A1}):] There exists $\mathcal{K}_{\mathcal{H}} >0$ such that $\mathcal{H}$ is a subset of  $ \Lambda_{1,\mathcal{K}_{\mathcal{H}}}(\mathcal{X},\mathcal{Y})$ and   $ \sup_{h \in \mathcal{H}} \|h\|_\infty < \infty$. 
 %
% \item [(\textbf{A1}):] There exists $\mathcal{K}_{\sigma} >0$ such that $\sigma \in \Lambda_{1,\mathcal{K}_{\sigma}}(\R) $.
 %
 \item [(\textbf{A1}):] $\mathcal{X} \subset \R^{\dx}$ is a compact set. \\
                        Under (\textbf{A1}), we set 
      \begin{equation} \label{def_R}
          R:=\max(1, 4\nm{\mx}).             
   \end{equation}                       
  \item [(\textbf{A2}):] There exists $\mathcal{K}_{\ell} >0$ such that, $ \ell \in \Lambda_{1,\mathcal{K}_{\ell}}(\R^{d_y} \times \mathcal{Y})$.  
  \end{enumerate}   
%

%\medskip

 For all $L, N, S, B, F > 0$, we set
 \begin{equation}\label{def_M_G}
 M_{L, N, S, B, F} := \max\Big(\sup_{h \in \mathcal{H}_{\sigma,\dx,\dy} } \sup_{z \in \mathcal{Z}} \ell(h,z) , 1 \Big) \text{ and }  G_{L, N, S, B, F} := \sup_{h_1, h_2 \in \mathcal{H}_{\sigma,\dx,\dy}, h_1 \neq h_2 } \sup_{ z\in \mathcal{Z}  } \frac{ |\ell(h_1,z) - \ell(h_2,z) | }{ \|h_1 - h_2 \|_\infty },
\end{equation}  
with $\mathcal{H}_{\sigma,\dx,\dy} = \mathcal{H}_{\sigma,\dx,\dy} (L, N, S, B, F)$.
When it is clear in the context and no confusion can arise, we use $M, G$ for $M_{L, N, S, B, F}$ and $G_{L, N, S, B, F}$ respectively. 
Under the assumption (\textbf{A1}) and if $\my$ is bounded, we have,
 \begin{equation}\label{eq_M_tilde_F}
 \widetilde{M}_F := \sup_{(y_1,y_2) \in B\big(\max(F,\|\my\|) \big)} \ell(y_1,y_2)<\infty,
\end{equation}
where $B\big(\max(F,\|\my\|) \big) \subset \R^\dy \times \R^\dy$ is the ball of radius $\max(F,\|\my\|)$, centered at 0. 
Let $z_0 =(x_0,y_0)$ fixed.
In addition to (\textbf{A2}), we have for all $z=(x,y) \in \mz$ and $h \in \mathcal{H}_{\sigma,\dx,\dy}$,
 \begin{align}\label{eq_ell_z0}
\nonumber \ell(h,z) &\leq  \ell(h,z_0)  + |\ell(h,z) - \ell(h,z_0) | \leq \ell( h(x_0), y_0) + |\ell( h(x), y) - \ell( h(x_0), y_0) | \\
  &\leq \widetilde{M}_F + \mk_\ell ( \|x-x_0 \| + \| y- y_0 \|) 
  \leq \widetilde{M}_F + 2 \mk_\ell ( \|\mx \| + \| \my\|).  
\end{align}
In this case, we get, 
 \begin{equation}\label{eq_Mlnsbf_M_tilde_F}
 M_{L, N, S, B, F} \leq \max\big(\widetilde{M}_F + 2 \mk_\ell ( \|\mx \| + \| \my\|), 1  \big) <\infty,
\end{equation}
 and this bound does not depend on $L, N, S, B$.
Also, under (\textbf{A2}), we get for all $h_1, h_2 \in \mathcal{H}_{\sigma,\dx,\dy}$ with $h_1 \neq h_2$  and $z=(x,y) \in \mz$ such that $h_1(x) \neq h_2(x)$,
 \begin{equation}\label{eq_Glnsbf}
\frac{ |\ell(h_1,z) - \ell(h_2,z) | }{ \|h_1 - h_2 \|_\infty } 
= \frac{ |\ell(h_1(x),y) - \ell(h_2(x),y) | }{ \|(h_1(x),y) - (h_2(x),y) \| } \times   \frac{\|(h_1(x),y) - (h_2(x),y) \|}{ \|h_1 - h_2 \|_\infty} \leq \mk_\ell.
\end{equation}
Hence,  $G_{L, N, S, B, F} \leq \mk_\ell$ and this bound does not depend on $L, N, S, B, F$.

\medskip
\noindent
Let us give the definition of the weak dependence in a general context, see \cite{doukhan1999new} and \cite{dedecker2007weak}.
Let $E$ be a separable Banach space.  
%For a set $E$, %$$h:E \rightarrow \R$, 
%we denote by $\Lambda(E)$  the set of functions $h:E \rightarrow \R$ such that $\lip(h)<\infty$ and 
%$\Lambda^{(1)} = \left\{h \in \Lambda,~~ \|h\|_{\infty}< \infty,~\lip(h) < 1 \right\}$, where
%\[
%\lip(h) =\sup \left\{ \frac{|h(x) - h(y)|}{\|x-y\|},~x, y \in E, ~ x\neq y\right\}.
%\]
%
\begin{Def}\label{def_weak_dep}
An $E$-valued process $(Z_t)_{t \in \Z}$ is said to be $(\Lambda_1(E),\psi,\epsilon)$-weakly dependent if there exists a function 
$\psi: [0,\infty)^2 \times \N^2 \rightarrow [0,\infty)$ and a sequence $\epsilon=(\epsilon(r))_{r \in \N}$ decreasing
to zero at infinity such that, 
for any $g_1,g_2 \in \Lambda_1(E)$ with $g_1: E^{u} \rightarrow \R$, $g_2:E^{v}  \rightarrow \R$ ($u, v \in \N$)
 and for any $u$-tuple $(s_1,\cdots,s_u)$ and any $v$-tuple $(t_1,\cdots,t_v)$ with $s_1 \leq \cdots \leq s_u \leq s_u +r \leq t_1 \leq \cdots \leq t_v$,
 the following inequality is fulfilled:
 \begin{equation*}%\label{eq_weak_dep}
 \left|\cov \left(g_1(Z_{s_1},\cdots,Z_{s_u}), g_2(Z_{t_1},\cdots,Z_{t_v})\right) \right| \leq \psi \left(\lip(g_1),\lip(g_2),u,v \right)\epsilon(r).
 \end{equation*}
\end{Def}
 The following choices of $\psi$ (see also \cite{dedecker2007weak}) are well known examples.
 \begin{itemize}
 \item $\psi \left(\lip(g_1),\lip(g_2),u,v \right)= v \lip(g_2)$: the $\theta$-weak dependence, then denote $\epsilon(r) = \theta(r)$;
 \item $\psi \left(\lip(g_1),\lip(g_2),u,v \right)= u \lip(g_1) + v \lip(g_2)$: the $\eta$-weak dependence, then denote $\epsilon(r) = \eta(r)$;
 \item $\psi \left(\lip(g_1),\lip(g_2),u,v \right)= u v \lip(g_1) \cdot \lip(g_2)$: the $\kappa$-weak dependence, then denote $\epsilon(r) = \kappa(r)$;
  \item $\psi \left(\lip(g_1),\lip(g_2),u,v \right)= u \lip(g_1) + v \lip(g_2) + u v \lip(g_1) \cdot \lip(g_2)$: the $\lambda$-weak dependence, then denote $\epsilon(r) = \lambda(r)$.   
 \end{itemize}
 %
% It is easy to see that $\eta(r) \leq \theta(r)$ for all $r \geq 0$. 
 %
% In the sequel, for each of the four choices of $\psi$ above, we set respectively,
% $\Psi(u,v)=2v$, $\Psi(u,v)=u+v$, $\Psi(u,v)=uv$ and $\Psi(u,v)=(u+v + uv)/2$. 
 
\medskip

Let us set now the weak dependence assumption.
\begin{enumerate}
 \item [(\textbf{A3}):]  The process $\{Z_t=(X_t,Y_t), ~ t \in \Z \}$ is stationary ergodic and $(\Lambda_1(\mathcal{Z}),\psi,\epsilon)$-weakly dependent with $\epsilon_r = \mathcal{O}(r^{-\gamma})$ for some $\gamma >3$.
\end{enumerate}
Numerous classical models satisfy (\textbf{A3}); for example, ARMAX, TARX, GARCH-X, ARMAX-GARCH, APARCH-X (\cite{francq2019qml}), multivariate INGARCH (with exponential family conditional distribution), see for instance \cite{diop2022statistical}, \cite{diop2022general}.
 Other examples such as APARCH-X$(\delta,\infty)$, ARX($\infty$)-ARCH($\infty$) introduced by \cite{diop2022inference} also satisfy (\textbf{A3}), see \cite{diop2022statistical}.

\section{H{\"o}lder smooth functions approximation by DNNs}
%
%\subsection{H{\"o}lder functions approximation by DNNs}
%
Let $U \subseteq \R^{d_x}$.
For any $\beta=(\beta_1,\cdots,\beta_{d_x})^T \in \N^{d_x}$ and $x=(x_1,\cdots,x_{d_x})^T \in U$, we set
\[ |\beta|=\sum_{i=i}^\dx \beta_i \text{ and } \partial^\beta = \dfrac{\partial^{|\beta|}}{\partial x_1^{\beta_1} \cdots \partial x_{d_x}^{\beta_{d_x}}}  .\]
For any $s>0$, the H{\"o}lder space $\mathcal{C}^s(U)$ is a set of vector valued functions $h: U \rightarrow \R^\dy$ such that, for any $\beta \in \N^\dx$ with $|\beta| \leq [s]$, $\| \partial^\beta h \|_\infty < \infty$ and for any  $\beta \in \N^\dx$ with $|\beta| = [s]$, $Lip_{s-[s]}(\partial^\beta h) < \infty$.
 This space is equipped with the norm
 \begin{equation*}
 \|h \|_{\mathcal{C}^s(U)} = \sum_{0\leq |\beta| \leq [s]} \| \partial^\beta h \|_\infty + \sum_{|\beta| = [s]} Lip_{s-[s]}(\partial^\beta h).
\end{equation*} 
For $s>0$, $U \subseteq \R^{d_x}$, $\mathcal{C}^s(\overline{U})$ with the norm $\|\cdot \|_{\mathcal{C}^s(\overline{U})}$ is a Banach space (see for example \cite{Triebel1992}, \cite{ern2004theory}).
For any $s>0$, $U \subseteq \R^{d_x}$ and $\mathcal{K}>0$, set
\begin{equation*}
 \mathcal{C}^{s,\mathcal{K}}(U) = \big\{h \in \mathcal{C}^s(U), ~  \| h\|_{\mathcal{C}^s(U)} \leq  \mathcal{K} \big\}.
\end{equation*}

\medskip

Let us consider the following definition, see also \cite{ohn2019smooth}, \cite{ohn2022nonconvex}.

\begin{Def}\label{def_pwl_quad}
Let a function $g: \R \rightarrow \R$.
\begin{enumerate}
\item $g$ is continuous piecewise linear (or "piecewise linear" for notational simplicity) if it is continuous and there exists $K$ ($K\in \N$) break points $a_1,\cdots, a_K \in \R$ with $a_1 \leq a_2\leq\cdots \leq a_K $ such that, for any $k=1,\cdots,K$, $g'(a_k-) \neq g'(a_k+)$ and $g$ is linear on $(-\infty,a_1], [a_1,a_2],\cdots [a_K,\infty)$.
\item $g$ is locally quadratic if there exits an interval $(a,b)$ on which $g$ is three times continuously differentiable with bounded derivatives and there exists $t \in (a,b)$ such that $g'(t) \neq 0$ and $g''(t) \neq 0$.
\end{enumerate} 
\end{Def}
The classical activation functions $\sigma(z) = \max(z,0)$ (ReLU) is piecewise linear and $\sigma(z) = 1/(1+e^{-z})$ (sigmoid) is locally quadratic (see \cite{ohn2019smooth}).  
 Consider the following assumption on the activation function $\sigma$.
 \begin{enumerate} 
 \item [(\textbf{A4}):] There exists $\mathcal{K}_{\sigma} >0$ such that $\sigma \in \Lambda_{1,\mathcal{K}_{\sigma}}(\R) $. Moreover, $\sigma$ is  either piecewise linear or locally quadratic and fixes a segment $I \subseteq [0,1]$.
  \end{enumerate}
\medskip

Recall that, $\sigma$ fixes the segment $I$ if $\sigma(z) = z$ for all $z \in I$.
Here are some examples of activation functions satisfying (\textbf{A4}).
\begin{itemize}
\item ReLU: $\sigma(z) = \max(z,0)$.
\item Leaky ReLU: $\sigma(z) = \max(z,az)$, for $a \in (0,1)$. 
\item Exponential linear unit (ELU) \cite{clevert2015fast}: $\sigma(z)=a(e^z -1)\ind(z \leq 0) + z \ind(z >0)$, for $a >0$.
\item Inverse square root linear unit (ISRLU) \cite{carlile2017improving}: $\sigma(z) = \frac{z}{\sqrt{1+az^2}}\ind(z \leq 0) + z \ind(z >0)$, for $a >0$. 
\item SignReLu \cite{lin2018research}:  $\sigma(z) = a\frac{z}{1-z} \ind(z < 0) + z \ind(z \geq 0) $, for $a \in (0,1)$. 
\end{itemize}

\medskip

 The following theorem provides an approximation of any H{\"o}lder smooth function by DNNs with a large class of activation functions.
\begin{thm}\label{prop_approx}
 Assume that (\textbf{A1}) and (\textbf{A4}) hold. 
Let $s, \mathcal{K}>0$ and $h: \mathcal{X} \rightarrow \R^\dy$ with $h \in \mathcal{C}^{s,\mathcal{K}}(\mathcal{X})$. There exist $L_0, N_0, S_0, B_0 > 0$ depending only on $d_x, d_y, s, \mathcal{K}$, $R$ (defined at (\ref{def_R})) and $\sigma$ such that, for all $\epsilon >0$, there exists a neural network $h_\epsilon \in  \mathcal{H}_{\sigma,\dx,\dy} (L_0 \log_+(1/\epsilon), N_0 \epsilon^{-\dx/s}, S_0 \epsilon^{-\dx/s} \log_+(1/\epsilon), B_0 \epsilon^{-4(d_x/s +1})$ satisfying,
 \begin{equation}\label{eq_prop_approx}
  \| h - h_\epsilon\|_{\infty,\mathcal{X}} \leq \epsilon,
\end{equation}  
where $\log_+ x = \max(1,\log x)$ for all $x>0$.
\end{thm}
Theorem \ref{prop_approx} is an extension of Theorem 1 of \cite{ohn2019smooth} to vector valued functions defined on any compact subset of $\R^\dx$. This theorem also extends Theorem 1 of \cite{schmidt2019deep} to deep neural networks with a broad class of activation functions.

\section{Excess risk bound}
Consider the learning problem presented in the introduction, based on the observations $D_n=\{Z_1=(X_1,Y_1),\cdots, Z_n=(X_n,Y_n) \}$ which is a trajectory  of a stationary and ergodic process $\{Z_t=(X_t,Y_t), ~ t \in \Z \}$, taking values in $ \mathcal{Z} = \mathcal{X} \times \mathcal{Y}$, with $\mathcal{Y} \subset \R$, that is, $d_y=1$. 
We focus on the class of DNNs $\mathcal{H}_{\sigma,\dx,1} (L_n,N_n, S_n, B_n, F_n)$ (defined at (\ref{def_H_lnsbf})), for some $L_n, N_n, S_n, B_n, F_n >0$.
Set for the sequel, $\mathcal{H}_{\sigma,n} := \mathcal{H}_{\sigma,\dx,1} (L_n,N_n, S_n, B_n, F_n)$.
Also, denote,
\begin{equation}\label{def_Mn_Gn} 
M_n:= M_{L_n, N_n, S_n, B_n, F_n}, ~ G_n:= G_{L_n, N_n, S_n, B_n, F_n}  \text{ and }  \widetilde{M}_n:= M_{F_n},
\end{equation} 
where $M_{L_n, N_n, S_n, B_n, F_n}$, $G_{L_n, N_n, S_n, B_n, F_n}$  and $M_{F_n}$ are defined in (\ref{def_M_G}) and (\ref{eq_M_tilde_F}).
The neural network obtained from the empirical risk minimization (ERM) algorithm is given by,
\begin{equation}\label{def_ERM_DNN}
    \widehat{h}_{n} = \underset{h\in \mathcal{H}_{\sigma,n}}{\argmin }\widehat{R}_{n}(h),
\end{equation}
where $\widehat{R}_{n}$ is defined at (\ref{def_emp_risk}).  
The excess risk is given by 
$ R(\widehat{h}_{n}) - \inf_{ h \in \mathcal{F}(\mathcal{X}, \mathcal{Y})} R(h) $, where $R$ is defined at (\ref{def_risk}).
This excess risk has the following well known decomposition,
\begin{equation}\label{excess_risk_decomp}
R(\widehat{h}_{n}) - \underset{h \in \mathcal{F}(\mathcal{X}, \mathcal{Y})}{\inf} R(h) = \underset{\text{Estimation error}}{ \underbrace{R(\widehat{h}_{n})  -  \underset{h \in \mathcal{H}_{\sigma,n}  }{\inf} R(h)} }     
 +  
 \underset{\text{Approximation error}}{ \underbrace{\underset{h \in \mathcal{H}_{\sigma,n}  }{\inf} R(h)  - \underset{h \in \mathcal{F}(\mathcal{X}, \mathcal{Y})}{\inf} R(h)}  }.
\end{equation}
In the sequel, we consider the target neural network $h_{\mathcal{H}_{\sigma,n}} \in \mathcal{H}_{\sigma,n}$, assumed to exist and satisfies,
\begin{equation}\label{cond_target_DNN}
  R( h_{\mathcal{H}_{\sigma,n}} ) =  \underset{h \in  \mathcal{H}_{\sigma,n}}{\inf} R(h) .
 \end{equation}

A major concern in statistical learning theory is to solve the problem of the best trade-off between the estimation and approximation errors.
This is done by taking into account the complexity of the class of hypothesis functions, which must not be too large, nor too small.

\medskip

In this section, we focus on the convergence rate of the excess risk, and set an additional assumption.

 \begin{enumerate} 
 \item [(\textbf{A5}):] There exists a predictor $h^* \in \mathcal{F}(\mathcal{X}, \mathcal{Y})$, satisfying $R(h^*) = \inf_{ h \in \mathcal{F}(\mathcal{X}, \mathcal{Y})} R(h)$ and $h^* \in \mathcal{C}^{s,\mathcal{K}}(\mathcal{X})$ for some $s, \mk >0$.
  \end{enumerate} 
The following theorem provides a bound of the excess risk.

\begin{thm}\label{theo_excess_risk}
Assume that (\textbf{A1})-(\textbf{A5}) hold. Let $\eta, \nu \in (0,1)$ and $\alpha > 2+ d_x/s$.
Then, there exist universal constants  $L_0, N_0, S_0, B_0 >0$ and
$n_0 \geq 1$, depending on $\mk_\ell, C_\sigma, d_x, \eta,\nu,\alpha, s, L_0, N_0, S_0, B_0$, such that, for any $n\geq n_0$, $F_n >0$, if for the DNN's class 
\[\mathcal{H}_{\sigma,n} = \mathcal{H}_{\sigma,\dx,1} \Big((L_0/\alpha) \log n , N_0n^{d_x / (s \alpha)}, (S_0/\alpha) n^{d_x / (s \alpha)} \log n , n^{4(d_x/s + 1)/\alpha}, F_n \Big),\]
 $M_n < \infty$, then,
we have with  probability at least $1-\eta$,
\begin{equation}\label{theo_eq_excess_risk}
R(\widehat{h}_{n}) - \underset{h \in \mathcal{F}(\mathcal{X}, \mathcal{Y})}{\inf} R(h) \leq \dfrac{2M_n + \mk_\ell}{n^{1/\alpha}} + \Bigg(\frac{\log(2C_1\log n/ \eta)}{C_{n,2}}\Bigg)^{1/2} 
 \end{equation}
for some constant $C_1 >0$, where $\widehat{h}_{n}$ is defined in (\ref{def_ERM_DNN}), $M_n$ is given in (\ref{def_Mn_Gn}) and (\ref{def_M_G}), and $C_{n,2}$ is given in (\ref{def_Cn2}).
\end{thm}

\begin{rmrk}\label{rmk1}
If $\my$ is bounded, and by choosing a constant sequence $F_n = F >0$, then, from (\ref{eq_Mlnsbf_M_tilde_F}), (\ref{def_Mn_Gn}) and (\ref{theo_eq_excess_risk}), we get,
\begin{equation}\label{rmk_eq_excess_risk}
R(\widehat{h}_{n}) - \underset{h \in \mathcal{F}(\mathcal{X}, \mathcal{Y})}{\inf} R(h) \leq \dfrac{2\max\big(\widetilde{M}_F + 2 \mk_\ell ( \|\mx \| + \| \my\|), 1  \big) + \mk_\ell}{n^{1/\alpha}} + \Bigg(\frac{\log(2C_1\log n/ \eta)}{C_{n,2}}\Bigg)^{1/2}. 
 \end{equation}
 Which shows that, the convergence rate of the excess risk is less than $\mathcal{O}(n^{-1/\alpha})$ for all $\alpha > 2+ d_x/s$. 
 This rate is close to $\mathcal{O}(n^{-1/2})$ when $s \gg d_x$.
 Note that, the rate obtained in \cite{ohn2019smooth} is close to $\mathcal{O}(1/n)$ when $s \gg d_x$. But, these findings are in specific cases of regression, classification, independent observations and with an input space $\mx = [0,1]^\dx$.
 The results obtained here are in a general setting and for a large class of dependent processes.
\end{rmrk}

%%%%%%%%%%%%%%%%%%%%%%%%%%%%%%%%%%%%%%%%%%%%%%%%%%%%%%%%%%%%%%%%%%%%%%%%%%
%%%%%%%%%%%%%%%%%%%%%%%%%%%%%%%%%%%%%%%%%%%%%%%%%%%%%%%%%%%%%%%%%%%%%%%%%%

\section{Proofs of the main results} 
 \subsection{Proof of Theorem \ref{prop_approx}}
 First of all, remark the parallelization property for two networks of the same depth: for any $d_1, d_2 \in \N$ and $L, N_1, N_2, S_1, S_2, B_1, B_2 >0$,
 \[ h_1 \in \mathcal{H}_{\sigma,\dx, d_1} (L, N_1, S_1, B_1), ~ h_2 \in \mathcal{H}_{\sigma,\dx, d_2} (L, N_2, S_2, B_2) \Rightarrow (h_1, h_2) \in \mathcal{H}_{\sigma,\dx, d_1 + d_2} (L, N_1+ N_2, S_1 + S_2, B_1+ B_2) .\] 
Thus, it suffices to prove the theorem with $d_y = 1$.
To do this, we split the proof into two parts, as in the proof of Theorem 1 in \cite{schmidt2019deep}.
Let $s, \mathcal{K}, \epsilon>0$ and $h \in \mathcal{C}^{s,\mathcal{K}}(\mathcal{X})$.  Without loss of generality, we assume that $\epsilon < 1$ and that, $\sigma$ fixes the segment $I=[1/4, 3/4]$. 

\medskip

\indent (1) Assume that $\mathcal{X} \subset [1/4, 3/4]^\dx$. \\
  \indent (i) Case of piecewise linear activation functions.\\
  Let $\rho$ be the ReLU activation function.
  From Theorem 1 in \cite{schmidt2019deep}, for any $m\geq 1$, $M \geq \max\{ 5^\dx, (s+1)^\dx, (\mathcal{K}+1)e^\dx \}$,  
  there exists a network $\widetilde{h}_1 \in  \mathcal{H}_{\rho,\dx, 1} (L_1, N_1, S_1, 1) $ with $L_1= 9 + (m+5)(1+ \lceil \log_2 \max(d_x, s) \rceil)$, $N_1=6(d_x + \lceil s \rceil) M$ and $S_1\leq 142 (d_x + s + 1)^{3+d_x} M (m + 6)$, such that,
  % for $R= \max(1, \| \mathcal{X}\|)$, % R=1 in this case
%
\begin{equation} \label{diff_h1_tilde_h}
\nm{\widetilde{h}_1 - h}_{\infty,\mathcal{X}} \leq (2 \mathcal{K} R^s + 1)(1+ d_x^2 + s^2)6^{d_x} M 2^{-m} + \mathcal{K} (9R)^s M^{-s/d_x}.
\end{equation} 
From Lemma A1 of \cite{ohn2019smooth}, there exists a network, with the activation function $\sigma$, that produces the same output of this ReLU neural network; that is, there exists $\widetilde{h}_2 \in \mathcal{H}_{\sigma,\dx, 1} (L_2, N_2, S_2, B_2) $ satisfying,
\begin{equation*}
\nm{\widetilde{h}_1 - \widetilde{h}_2}_{\infty,\mathcal{X}} = 0,
\end{equation*} 
with $L_2=L_1$, $N_2 = 2N_1$, $S_2=4S_1 + 2L_1 N_1 + 1$ and for some $B_2 >0$  depending on $\sigma$.
Therefore, we get from (\ref{diff_h1_tilde_h}) 
\begin{equation} \label{diff_h2_tilde_h}
\nm{h - \widetilde{h}_2}_{\infty,\mathcal{X}} \leq (2 \mathcal{K} R^s + 1)(1+ d_x^2 + s^2)6^{d_x} M 2^{-m} + \mathcal{K} (9R)^s M^{-s/d_x}.
\end{equation} 
By taking $M=\max\Big\{5^\dx, (s+1)^\dx, (\mathcal{K}+1)e^\dx, (9R)^\dx (2 \mk)^{d_x/s} \epsilon^{-d_x/s} \Big\}$ and $m = \max \Big \{ \log_2 \Big( 2(2 \mk R^s +1)(1+d_x^2 +s^2) 6^\dx M \epsilon^{-1}\Big) , 1 \Big \}$ in (\ref{diff_h2_tilde_h}), we get 
\begin{equation} \label{diff_h2_tilde_h_ep}
\nm{h - \widetilde{h}_2}_{\infty,\mathcal{X}} \leq \epsilon.
\end{equation} 
For these choices of $M$ and $m$, we have,
$M\leq 5^\dx + (s+1)^\dx + (\mathcal{K}+1)e^\dx + 9^\dx (2 \mk)^{d_x/s} \epsilon^{-d_x/s}  $ and $ m \leq 1 + \log_2 \Big( 2(\mk +1)(1+d_x^2 +s^2) 6^\dx M \epsilon^{-1}\Big)  $. Hence, one can easily see that:
 \[ L_2 = L_1= 9 + (m+5)(1+ \lceil \log_2 \max(d_x, s) \rceil) \leq L_0 \log (1/\epsilon), ~ N_2 = 2N_1 = 12 (d_x + \lceil s \rceil) M \leq N_0 \epsilon^{-d_x/s} , \]
 \[S_2 = 4S_1 + 2L_1 N_1 + 1 \leq 568 (d_x + s + 1)^{3+d_x} M (m + 6) + 2 L_0 \big(\log (1/\epsilon) \big) N_0 \epsilon^{-d_x/s} + 1 \leq S_0 \epsilon^{-d_x/s} \log (1/\epsilon) , \]
 and  $B_2 \leq B_0 \epsilon^{-4(d_x/s +1)}$  for some $L_0, N_0, S_0, B_0 >0$ depending on $d_x, s, \mk, R$ and $\sigma$.
Thus, the result holds for piecewise linear activation functions and when $\mathcal{X} \subset [1/4, 3/4]^\dx$.

\medskip

(ii) Case of locally quadratic activation functions. \\
   Let $\mathfrak{M} \in \N$. Consider the grid points with length $1/\mathfrak{M}$ inside the $d_x$-dimensional unit hypercube $[0,1]^\dx$,
   \begin{equation}\label{def_G_dx_M}
   \mathbb{G}_{d_x,\mathfrak{M}} := \Big \{\dfrac{1}{\mathfrak{M}} (m_1,\cdots,m_\dx), ~ m_j \in \{0,1,\cdots \mathfrak{M} \}, j=1,\cdots,d_x   \Big \}. 
   \end{equation}
For all $ z \in \mx \cap \mathbb{G}_{d_x,\mathfrak{M}} $, denote by $P^s_{z,\mathfrak{M}}h(\cdot)$ the $\lfloor s \rfloor$-th order  Taylor polynomial of $h$ around $\pmb{z}$, that is
   \begin{equation}\label{def_Pzh_1}
  P^s_{z}h(x)= \sum_{\beta \in \N_0^\dx, |\beta| \leq s } (\partial f)(z) \dfrac{(x-z)^\beta}{\beta!}.
   \end{equation} 
When $z \in [0,1]^\dx \setminus \mx$, set $ z^* = \underset{\widetilde{z} \in \argmin_{u \in  \mx \cap \mathbb{G}_{d_x,\mathfrak{M}} }(\| z - u \|) }{\argmin} \nm{\widetilde{z}} $. One can easily check that, $ \pmb{z}^*$ exists and is unique. Therefore, set
 \begin{equation}\label{def_Pzh_2} 
 P^s_{z}h(x) = P^s_{z^*}h(x).
\end{equation} 
Define for all $x=(x_1,\ldots,x_\dx)', z=(z_1,\ldots,z_\dx)' \in \mathbb{G}_{d_x,\mathfrak{M}}$,
 \begin{equation}\label{def_Ph} 
 P^s h(x) = \sum_{z \in \mathbb{G}_{d_x,\mathfrak{M}}} P^s_{z}h(x) \prod_{j=1}^\dx \big( 1 - \mathfrak{M} |x_j - z_j| \big)_+ .
\end{equation}
For $\mathfrak{M} \geq 4$, we get from the proof of Theorem 1 in \cite{schmidt2019deep},
 \begin{equation}\label{diff_Ph_h_norm_inf} 
 \nm{ P^s h - h }_{\infty, \mx} \leq \mk 3^s \mathfrak{M}^{-s}. 
\end{equation}
According to the proof of Theorem 1 in \cite{ohn2019smooth}, for some $L_3, N_3, S_3, B_3 > 0$, one can find a neural network,

\begin{equation*} 
\widetilde{h}_3 \in \mathcal{H}_{\sigma,\dx, 1} \Big(L_3 \log(1/\epsilon), N_3 \epsilon^{-d_x/s}, S_3 \epsilon^{-d_x/s} \log(1/\epsilon), B_3 \epsilon^{-4(d_x/s + 1)} \Big) 
\end{equation*} 
such that,
\begin{equation}\label{diff_Ph_h_h3}
\nm{P^s h - \widetilde{h}_3}_{\infty,\mathcal{X}} \leq C_1 \epsilon ~ \text{ for some } C_1 >0.
\end{equation} 
By taking $\mathfrak{M} = [\epsilon^{-1/s}]$, the result follows from (\ref{diff_Ph_h_norm_inf}) and (\ref{diff_Ph_h_h3}).

\medskip

\indent (2) The general case of any compact set $\mx$. \\
  As in the proof of Theorem 1 in \cite{schmidt2019deep}, define the affine map, 
 \begin{equation}\label{def_map_T}  
  T: \R^{d_x} \rightarrow \R^{d_x}, ~ x \mapsto x/R + (1/2,\ldots,1/2)^T,
  \end{equation}
where $R$ is defined in (\ref{def_R}). 
 By setting $\mx':=T(\mx)$, one can see that $\mx' \subseteq [1/4, 3/4]^{d_x}$.
 Also, one can easily verify that, since $h \in \mathcal{C}^{s,\mathcal{K}}(\mx)$, then, $g:=h(T^{-1}) \in \mathcal{C}^{s, R^s \mk}(\mx')$.
Therefore, let us apply the result of the first part to $g \in  \mathcal{C}^{s, R^s \mk}(\mx')$.  
So, for some $L_0, N_0, S_0, B_0 > 0$  depending only on $d_x, s, \mathcal{K}$ and $R$, there exists a neural network $\widetilde{g} \in  \mathcal{H}_{\sigma,\dx,1} (L_0 \log_+(1/\epsilon), N_0 \epsilon^{-\dx/s}, S_0 \epsilon^{-\dx/s} \log_+(1/\epsilon), B_0 \epsilon^{-4(d_x/s +1)})$ satisfying,
 \begin{equation}\label{inq_g_tilde}
  \| g - \widetilde{g} \|_{\infty,\mx'} \leq \epsilon.
\end{equation}  
Now, consider the neural network,

 \begin{equation}\label{def_h_tilde_part2}
 \widetilde{h} := \widetilde{g}\circ \sigma_0 \circ T , \text{ where } 
 \sigma_0(z_1\ldots,z_\dx)=\big(\sigma(z_1),\ldots,\sigma(z_\dx) \big) \text{ for all } (z_1\ldots,z_\dx) \in \R^\dx. 
\end{equation}  
Since $T$ is an affine transformation, the neural network $\widetilde{h}$ is obtained from $\widetilde{g}$ by adding one hidden layer, which also adds $2 \dx $ non-zero parameters.
That is,
\begin{equation}\label{h_tilde_part2_archit}
\widetilde{h} \in  \mathcal{H}_{\sigma,\dx,1} \Big( (L_0 +1) \log_+(1/\epsilon), N_0 \epsilon^{-\dx/s}, (S_0 + 2d_x) \epsilon^{-\dx/s} \log_+(1/\epsilon), \max(B_0,1) \epsilon^{-4(d_x/s +1)} \Big).
\end{equation}  
Also, $T(\mx) = \mx' \subseteq [1/4, 3/4]$ and, since this set is fixed by $\sigma$, we get from (\ref{inq_g_tilde}),
 \begin{equation}\label{h_h_tilde_diff_part2}
  \| h - \widetilde{h} \|_{\infty,\mx} = \| g\circ T - \widetilde{g}\circ \sigma_0 \circ T \|_{\infty,\mx} \leq \| g- \widetilde{g} \|_{\infty,\mx'} \leq \epsilon.
\end{equation}  
Thus, the result holds from (\ref{h_tilde_part2_archit}) and (\ref{h_h_tilde_diff_part2}). 
This completes the proof of the theorem.
 
 \qed

 \subsection{Proof of Theorem \ref{theo_excess_risk}}
 %
 % Let $F_n >0$ and a DNN's class $\mathcal{H}_{\sigma,n} \subset \mathcal{F}(\mathcal{X}, \mathcal{Y})$ such that $\|h\|_{\infty, \mx}\leq F_n$ for all $h \in \mathcal{H}_{\sigma,n}$.
 %
 Let $L_n,N_n, S_n, B_n, F_n >0$ and consider the DNN's class $\mathcal{H}_{\sigma,n} = \mathcal{H}_{\sigma,\dx,1} (L_n,N_n, S_n, B_n, F_n) \subset \mathcal{F}(\mathcal{X}, \mathcal{Y})$.
 Under (\ref{cond_target_DNN}), (\textbf{A5}) and the decomposition (\ref{excess_risk_decomp}), we get,
\begin{equation}\label{proof_ER_decomp}
R(\widehat{h}_{n}) - \underset{h \in \mathcal{F}(\mathcal{X}, \mathcal{Y})}{\inf} R(h) = R(\widehat{h}_{n}) -  R( h^* ) = \Big(  R(\widehat{h}_{n}) -  R( h_{\mathcal{H}_{\sigma,n}} ) \Big) + \Big(  R( h_{\mathcal{H}_{\sigma,n}} )  - R( h^* )\Big).
 \end{equation}
Let  $\eta, \nu \in (0,1)$ and $\alpha >2 + d_x/s$.
Let us deal first with the approximation error. 
Observe that, from the Lipschitz property of $\ell$, it holds for all $h \in \mathcal{H}_{\sigma,n}$, that,
\begin{equation*} %\label{proof_ER_eq_lip_prop_ell}
R( h )  - R( h^* ) = \E_{Z_0}\big[\ell \big( h(X_0),Y_0 \big) \big] -  \E_{Z_0}\big[\ell \big( h^*(X_0),Y_0 \big) \big]  \leq \mk_\ell \E_{X_0}| h(X_0) - h^*(X_0) |.
\end{equation*}
Hence, we get,
\begin{align}  \label{proof_ER_eq_AE}
\nonumber R( h_{\mathcal{H}_{\sigma,n}} )  - R( h^* ) &= \underset{h \in  \mathcal{H}_{\sigma,n}}{\inf} R(h) - R( h^* )  =  \underset{h \in  \mathcal{H}_{\sigma,n}}{\inf} \Big(R(h) - R( h^* ) \Big) \\
&\leq  \mk_\ell  \underset{h \in  \mathcal{H}_{\sigma,n}}{\inf} \Big( \E_{X_0} \big| h(X_0) - h^*(X_0) \big| \Big)
\end{align} 
Since $h^* \in \mathcal{C}^{s,\mathcal{K}}(\mathcal{X})$,
for $\epsilon = n^{-1/\alpha}$ in Theorem \ref{prop_approx}, one can find positive constants $L_0,N_0, S_0, B_0$ such that, with 
\begin{equation}\label{proof_ER_def_Ln_Nn_Sn_Bn}
L_n=\frac{L_0}{\alpha} \log n , N_n= N_0 n^{d_x / (s \alpha)} ,  S_n = \frac{S_0}{\alpha} n^{d_x / (s \alpha)} \log n \text{ and } B_n = B_0 n^{4(d_x/s + 1)/\alpha}
\end{equation}
there is a neural network $h_n \in \mathcal{H}_{\sigma,n} = \mathcal{H}_{\sigma,\dx,1} (L_n,N_n, S_n, B_n, F_n) $ satisfying,
\begin{equation*} %\label{proof_ER_diff_hn_h_star} 
\| h_n - h^*\|_{\infty, \mx} \leq \frac{1}{n^{1/\alpha}}.
\end{equation*}
Therefore, in addition to (\ref{proof_ER_eq_AE}), it holds that,
\begin{equation}\label{proof_ER_eq_AE_bound} 
R( h_{\mathcal{H}_{\sigma,n}} )  - R( h^* ) \leq  \frac{  \mk_\ell }{n^{1/\alpha}}.
\end{equation}
Let us consider now the estimation error over the class  $ \mathcal{H}_{\sigma,n} = \mathcal{H}_{\sigma,\dx,1} (L_n,N_n, S_n, B_n, F_n) $ where $L_n,N_n, S_n, B_n$ are given in (\ref{proof_ER_def_Ln_Nn_Sn_Bn}).
Under (\textbf{A1})-(\textbf{A3}), one can find a constant $C_{n,1} >0$ (see \cite{hwang2013study}) such that, for all $h \in \mathcal{H}_{\sigma,n}$,
 \begin{equation}\label{proof_ER_const_var}
 \mathbb{E}\Big[\Big(\sum_{i=1}^{n}\Big(\ell(h(X_{i}), Y_{i}) - \mathbb{E}[\ell(h(X_{0}), Y_{0})]\Big)\Big)^{2}\Big]\leq C_{n,1} n.
 \end{equation}
With the assumption $\alpha > 2+ d_x/s$, one can easily see that the condition (4.7) of Theorem 4.3 in \cite{kengne2023deep} holds.
From this theorem, there exists a constant $C_1 >0$ and $\widetilde{n}_0$ depending on $\mk_\ell, C_\sigma, d_x, \eta,\nu,\alpha, s$ such that, for any $n \geq \widetilde{n}_0$ and with  probability at least $1-\eta$, the estimation error satisfies,
\begin{equation}\label{proof_ER_eq_est_R}
 R(\widehat{h}_{n}) -  R( h_{\mathcal{H}_{\sigma,n}} ) \leq \dfrac{2M_n}{n^{1/\alpha}} + \Bigg(\frac{\log(2C_1\log n/ \eta)}{C_{n,2}}\Bigg)^{1/2},
\end{equation}
where,
\begin{equation}\label{def_Cn2}
 C_{n,2}=\frac{n^{2}}{nC_{n,1}+\log n \; n^{\nu-1/4}(2M_n)^{\nu}/C_{n,1}},
\end{equation}
with $C_{n,1}$ defined in (\ref{proof_ER_const_var}).  
Thus, the result follows from (\ref{proof_ER_decomp}), (\ref{proof_ER_eq_AE_bound}) and (\ref{proof_ER_eq_est_R}). This completes the proof of the theorem.

\qed


\begin{thebibliography}{10}

\bibitem{bauer2019deep}
{\sc Bauer, B., and Kohler, M.}
\newblock {On deep learning as a remedy for the curse of dimensionality in
  nonparametric regression}.
\newblock {\em The Annals of Statistics 47}, 4 (2019), 2261 -- 2285.

\bibitem{carlile2017improving}
{\sc Carlile, B., Delamarter, G., Kinney, P., Marti, A., and Whitney, B.}
\newblock Improving deep learning by inverse square root linear units (isrlus).
\newblock {\em arXiv preprint arXiv:1710.09967\/} (2017).

\bibitem{clevert2015fast}
{\sc Clevert, D.-A., Unterthiner, T., and Hochreiter, S.}
\newblock Fast and accurate deep network learning by exponential linear units
  (elus).
\newblock {\em arXiv preprint arXiv:1511.07289\/} (2015).

\bibitem{dedecker2007weak}
{\sc Dedecker, J., Doukhan, P., Lang, G., Le\'on, J.~R., Louhichi, S., and
  Prieur, C.}
\newblock {\em {Weak dependence: With Examples and Applications}}.
\newblock Lecture Notes in Statistics {\bf 190}, Springer-Verlag, New York,
  2007.

\bibitem{diop2022general}
{\sc Diop, M.~L., and Kengne, W.}
\newblock A general procedure for change-point detection in multivariate time
  series.
\newblock {\em TEST\/} (2022), 1--33.

\bibitem{diop2022inference}
{\sc Diop, M.~L., and Kengne, W.}
\newblock Inference and model selection in general causal time series with
  exogenous covariates.
\newblock {\em Electronic Journal of Statistics 16}, 1 (2022), 116--157.

\bibitem{diop2022statistical}
{\sc Diop, M.~L., and Kengne, W.}
\newblock Statistical learning for $\psi$-weakly dependent processes.
\newblock {\em arXiv preprint arXiv:2210.00088\/} (2022).

\bibitem{doukhan1999new}
{\sc Doukhan, P., and Louhichi, S.}
\newblock A new weak dependence condition and applications to moment
  inequalities.
\newblock {\em Stochastic processes and their applications 84}, 2 (1999),
  313--342.

\bibitem{ern2004theory}
{\sc Ern, A., and Guermond, J.-L.}
\newblock {\em Theory and practice of finite elements}, vol.~159.
\newblock Springer, 2004.

\bibitem{francq2019qml}
{\sc Francq, C., et~al.}
\newblock Qml inference for volatility models with covariates.
\newblock {\em Econometric Theory 35}, 1 (2019), 37--72.

\bibitem{hwang2013study}
{\sc Hwang, E., and Shin, D.~W.}
\newblock A study on moment inequalities under a weak dependence.
\newblock {\em Journal of the Korean Statistical Society 42}, 1 (2013),
  133--141.

\bibitem{kengne2023deep}
{\sc Kengne, W., and Modou, W.}
\newblock Deep learning for $\psi $-weakly dependent processes.
\newblock {\em arXiv preprint arXiv:2302.00333\/} (2023).

\bibitem{kim2021fast}
{\sc Kim, Y., Ohn, I., and Kim, D.}
\newblock Fast convergence rates of deep neural networks for classification.
\newblock {\em Neural Networks 138\/} (2021), 179--197.

\bibitem{kohler2020rate}
{\sc Kohler, M., and Krzyzak, A.}
\newblock On the rate of convergence of a deep recurrent neural network
  estimate in a regression problem with dependent data.
\newblock {\em arXiv preprint arXiv:2011.00328\/} (2020).

\bibitem{kurisu2022adaptive}
{\sc Kurisu, D., Fukami, R., and Koike, Y.}
\newblock Adaptive deep learning for nonparametric time series regression.
\newblock {\em arXiv preprint arXiv:2207.02546\/} (2022).

\bibitem{lin2018research}
{\sc Lin, G., and Shen, W.}
\newblock Research on convolutional neural network based on improved relu
  piecewise activation function.
\newblock {\em Procedia computer science 131\/} (2018), 977--984.

\bibitem{ma2022theoretical}
{\sc Ma, M., and Safikhani, A.}
\newblock Theoretical analysis of deep neural networks for temporally dependent
  observations.
\newblock {\em arXiv preprint arXiv:2210.11530\/} (2022).

\bibitem{ohn2019smooth}
{\sc Ohn, I., and Kim, Y.}
\newblock Smooth function approximation by deep neural networks with general
  activation functions.
\newblock {\em Entropy 21}, 7 (2019), 627.

\bibitem{ohn2022nonconvex}
{\sc Ohn, I., and Kim, Y.}
\newblock Nonconvex sparse regularization for deep neural networks and its
  optimality.
\newblock {\em Neural Computation 34}, 2 (2022), 476--517.

\bibitem{petersen2018optimal}
{\sc Petersen, P., and Voigtlaender, F.}
\newblock Optimal approximation of piecewise smooth functions using deep relu
  neural networks.
\newblock {\em Neural Networks 108\/} (2018), 296--330.

\bibitem{schmidt2019deep}
{\sc Schmidt-Hieber, J.}
\newblock Deep relu network approximation of functions on a manifold.
\newblock {\em arXiv preprint arXiv:1908.00695\/} (2019).

\bibitem{schmidt2020nonparametric}
{\sc Schmidt-Hieber, J.}
\newblock Nonparametric regression using deep neural networks with relu
  activation function.
\newblock {\em The Annals of Statistics\/} (2020), 1875 -- 1897.

\bibitem{suzuki2018adaptivity}
{\sc Suzuki, T.}
\newblock Adaptivity of deep relu network for learning in besov and mixed
  smooth besov spaces: optimal rate and curse of dimensionality.
\newblock {\em arXiv preprint arXiv:1810.08033\/} (2018).

\bibitem{Triebel1992}
{\sc Triebel, H.}
\newblock {\em Theory of function spaces {II}}, vol.~84 of {\em Monogr. Math.,
  Basel}.
\newblock Basel: Birkh{\"a}user Verlag, 1992.

\bibitem{yarotsky2017error}
{\sc Yarotsky, D.}
\newblock Error bounds for approximations with deep relu networks.
\newblock {\em Neural Networks 94\/} (2017), 103--114.

\end{thebibliography}
\end{document}